\documentclass[letterpaper, 10 pt, conference]{ieeeconf}
\IEEEoverridecommandlockouts  
\usepackage{url}

\usepackage{math}
\usepackage[all]{nowidow}

\usepackage{algpseudocode}
\usepackage{algorithm}

\usepackage{float}
\usepackage{graphicx}
\usepackage{cite}
\usepackage{booktabs} 
\long\def\invis#1{}

\begin{document}
%
\title{\LARGE \bf

Sonar–GPS Fusion for Seabed Mapping in Turbid Shallow Waters \\with an Autonomous Surface Vehicle
}

\author{Yisheng Zhang$^1$, Michael Xu$^{1}$, Alan Williams$^{2}$, Matthew Gray$^{2}$, Nare Karapetyan$^{3,\dagger}$, Miao Yu$^{1,\dagger}$
\thanks{This work was supported by the United States Department of Agriculture (USDA) National Institute of Food and Agriculture (NIFA) Sustainable Agricultural Systems (SAS) Program under Grant 2020-68012-31805 and in part by Independent Research \& Development Program (IR\&D) at Woods Hole Oceanographic Institution (WHOI).}
\thanks{$^1$University of Maryland,
College Park, MD 20742, USA, {\tt\small\{mixu1235, yiszhang, mmyu\}@umd.edu}}%
\thanks{$^2$University of Maryland Center for Environmental Science, Cambridge, MD 21613, USA, {\tt\small\{awilliams, mgray\}@umces.edu}}%
\thanks{$^3$Woods Hole Oceanographic Institution, Woods Hole, MA 02543, USA. {\tt\small nare@whoi.edu}}%
\thanks{$^\dagger$ Equal Contribution.}%
}

\maketitle
  
\begin{abstract}
Accurate seabed mapping is essential for habitat monitoring and infrastructure inspection. In turbid, shallow coastal waters, such as shellfish aquaculture farms, the effectiveness of traditional optical methods is limited. Autonomous surface vehicles (ASVs) equipped with forward-looking sonar (FLS) offer a promising alternative. However, existing sonar-based systems face challenges in achieving fine resolution mapping over long trajectories due to low-resolution positioning measurements and accumulated drift over long trajectories. In this paper, we present a drift-resilient seabed mapping framework that integrates local FLS frame alignment using the Fourier–Mellin transform (FMT) with global trajectory optimization based on an extended Kalman filter (EKF) that fuses global positioning system (GPS), inertial measurement unit (IMU), and compass data. A variance-based image blending strategy is used to further reduce visual artifacts in overlapping regions. Field trials on a structured oyster farm site show that our framework helps reduce drift in RMSE by 9.5\% relative to the FMT-only baseline. This framework also enables  sub-meter reconstruction accuracy and preservation of high-resolution textures needed for oyster inventory estimation within the mapped areas.

\end{abstract}

\invis{}

\section{INTRODUCTION}
\label{section:introduction}

Reliable seabed mapping is critical for a range of applications including  marine habitat monitoring, resource exploration, infrastructure inspection, and search-and-rescue missions. Traditional optical imaging methods, however, are severely constrained in low-visibility underwater environments due to high turbidity, low illumination, and strong light scattering. Under these conditions, forward-looking sonar (FLS) has been widely adopted due to its robustness. 
However, registration of overlapping sonar frames remains challenging due to low signal-to-noise ratio of sonar images, partial overlaps, and shadow cast artifacts~\cite{negahdaripour2013,hurtos2013}. Traditional registration approaches such as feature-based, template-based, region-based, and Fourier-based methods struggle to maintain robustness due to noisy sonar imagery and the simplified 2-D similarity transforms used in these methods ~\cite{aykin2013feature,hurtos2012,hurtos2015}. Furthermore,  although satisfactory local alignment can be achieved, due to the inevitably accumulated residual errors over long sequences, these methods suffer from increasingly pronounced drift in large-scale and/or long-range mapping tasks.

\begin{figure}[t]
    \centering
    {\includegraphics[width=0.98\columnwidth]{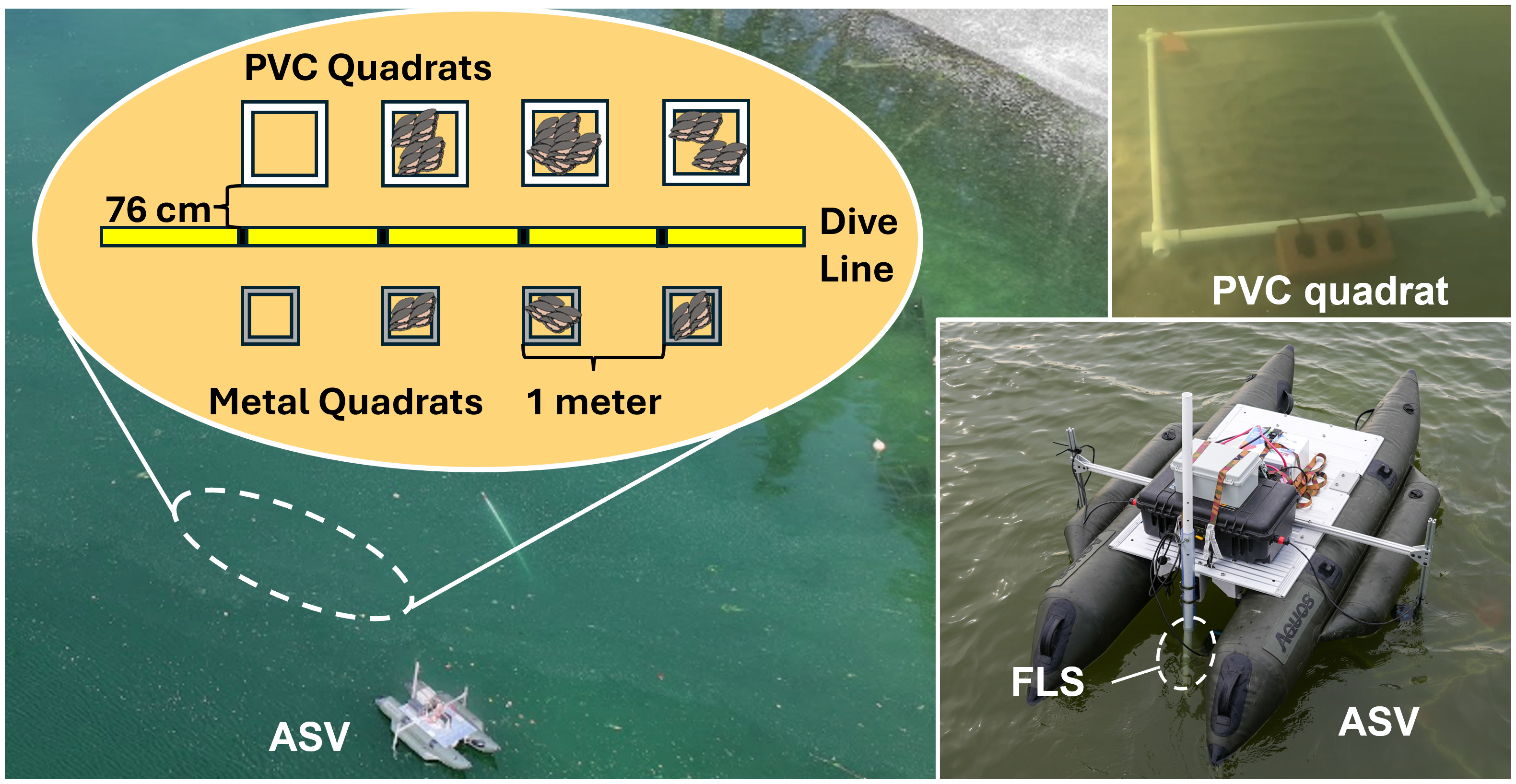}}
    \caption{Field experimental setup. An ASV equipped with a multi-beam FLS  was deployed to survey a farm site with pre-arranged on-bottom quadrats. Oysters are distributed within the quadrats. The top right inset shows a photo of a sample PVC quadrat from up close.}
    \label{fig:field_diagram}
\end{figure}

Autonomous surface vehicles (ASVs) equipped with a FLS and a global positioning system (GPS) offer a promising platform for seabed mapping\cite{karapetyan2019riverine}. While the on-board GPS can provide global references to correct sonar-based registration errors, it lacks the fine-grained accuracy required for high-resolution mapping. This gap is especially evident in mapping of on-bottom oyster farms in shallow, highly-turbid coastal waters, where surveying large areas of oyster inventory is needed to enable precision oyster farming.  The current inventory monitoring approaches rely on divers performing manual counting or dredge sampling.

Although individual oysters are hardly identifiable in FLS imagery, this work aims to generate accurate georeferenced seabed mosaics on a sub-meter scale and improve texture quality to support subsequent mapping of oyster inventory. These mosaics establish a spatial framework that enables the integration of machine-learning based classification algorithms to estimate oyster density and size within a mapped area~\cite{xu2025oyster}. Our work addresses a critical bottleneck: the lack of a low-cost, drift-resilient mapping baseline in turbid environments.

We propose a unified pipeline that integrates local sonar alignment, global trajectory optimization, and adaptive image blending. Our main contributions are summarized as follows.

\begin{itemize}
  \item \textbf{Drift-resilient registration:} We introduce a framework that fuses Fourier–Mellin transform (FMT)-based local alignment with global priors from an extended Kalman filter (EKF) (GPS, compass, and inertial
measurement unit (IMU)) through a global least-squares optimization. This enforces global consistency in sonar image registration and significantly mitigates cumulative drift, which is a key limitation of prior FMT-only approaches.
  
  \item \textbf{Detail-preserving robust blending:} We develop a variance-based blending strategy that combines exponential moving
averages (EMA)-smoothed short- and long-term variance statistics with adaptive weighted averaging. This enables reliable pixel fusion in overlapping sonar frames, effectively suppressing striping and noise while preserving fine seabed details—addressing the artifact-prone results of existing blending methods.
  
  \item \textbf{Field validation with low-cost ASVs:} We demonstrate the full pipeline in shallow, turbid coastal waters using a low-cost ASV platform (Fig.~\ref{fig:field_diagram}). The system achieves sub-meter accuracy with a 9.5\% drift RMSE reduction over the FMT-only baseline,  outperforms other representative registration methods, and delivers clearer, more detailed textures than existing blending strategies, demonstrating practical viability for habitat mapping.
\end{itemize}

\begin{figure*} [t]
    \centering
    \includegraphics[width=0.95
    \linewidth]{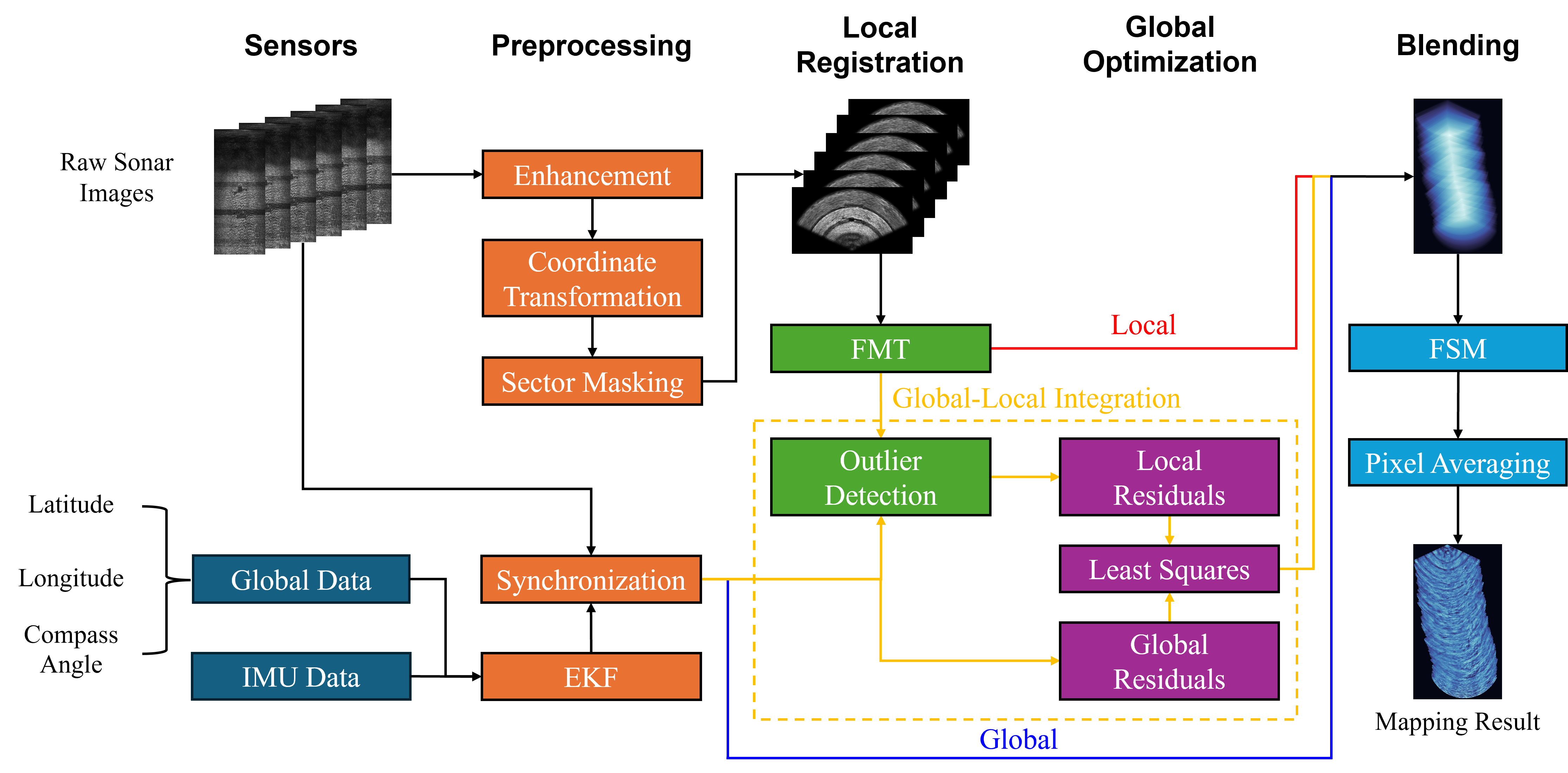} 
    \caption{The architecture of the proposed system. The pipeline integrates global priors and sonar images through preprocessing, local registration, global optimization, and blending to construct a globally consistent 2-D map. Preprocessing modules enhance sonar images before registration. Local alignment is performed with FMT, while global consistency is enforced by least-squares optimization. Variance-based pixel blending resolves conflicts in overlapping regions. In the diagram, \textbf{black arrows} denote shared preprocessing and blending steps; \textbf{yellow arrows and box} represent the joint local–global optimization pipeline (used as the single-stage scheme and as Stage~2 of the two-stage scheme); \textbf{red arrow} indicates Stage~1 of the two-stage scheme, where submaps are produced from local registration and blending without global correction; and \textbf{blue arrow} represents the for-reference scheme that builds maps solely from prior global estimates (without local registration).}
    \label{fig:workflow}
    \vspace{-3mm}
\end{figure*}

\section{Related Work}
\label{section:related_work}

Underwater mapping and navigation have been widely studied, spanning sonar image registration and sensor fusion with iterative optimization. While most prior work rely on autonomous underwater vehicles (AUVs), ASVs can directly access GPS and follow near-planar trajectories, making them well suited for constructing textured 2-D overhead mosaics. Many existing pipelines are rooted in simultaneous localization and mapping (SLAM), which emphasizes landmark tracking and sparse point-cloud mapping for navigation and 3-D reconstruction. Although these pipelines are effective for localization, they cannot be used to generate dense, interpretable 2-D seabed maps needed for ecological monitoring. Our evaluation emphasizes mosaicing fidelity and georeferencing consistency, which are not directly captured by SLAM-centric localization and loop-closure metrics.

\subsection{Sonar-Based Mapping and Sensor Fusion}

Early studies on FLS imagery focused on mosaicing through direct frame-to-frame registration. Hurtós et al.~\cite{hurtos2013} applied Fourier-based phase correlation, which was proven to be robust but was limited to 2-D similarity assumptions. To advance this method, Aykin et al.~\cite{aykin2013feature} proposed a feature-based method incorporating parallax and elevation cues. While these approaches achieve reliable local alignment, they lack mechanisms to suppress drift over long trajectories and often rely on high-frequency, costly sensors~\cite{hurtos2015,liu2025robust}, limiting their applicability for large-area mosaicing.

To mitigate drift, external references were incorporated. Li et al.~\cite{li2018realtime} proposed a real-time pipeline for a 3-D phased-array imaging sonar, initialized with GPS and aligned by using a variant of the iterative closest point (ICP) algorithm~\cite{besl1992method,zhang1994iterative}. While effective, this method resulted in significant computational overhead on large datasets. Dos Santos et al.~\cite{dosSantos2019fusion} combined sonar imagery with aerial photography, by applying ICP to register sonar with aerial maps to enhance localization in structured environments. While these methods demonstrated the benefits of external anchoring, they incurred increased sensing and computational requirements.

SLAM-based methods that integrate sonar with inertial and other onboard sensors were also explored and typically framed as pose-graph or factor-graph optimization~\cite{li2018pose}. Early work fused dense sonar features with Doppler velocity log (DVL)/IMU in a graph-based framework~\cite{johannsson2010imaging}, and iSAM2 enabled scalable incremental optimization~\cite{kaess2012isam2}. ICP was widely adopted in underwater SLAM as a fundamental alignment method~\cite{kim2023,wu2023}. Building on modality-specific advances, Cheung et al.~\cite{cheung2019} extended SLAM to synthetic aperture sonar with a non-Gaussian formulation, achieving long-mission robustness without expensive inertial aids. For multi-robot operations, DRACo-SLAM and its successor DRACo-SLAM2 represented sonar maps as object graphs and enabled efficient distributed loop closure~\cite{mcconnellDRACoSLAM,mcconnellDRACoSLAM2}. McConnell et al. explored cross-modal integration: one study employed FLS observations in a particle filter with the adaptive Monte Carlo localization (AMCL) against aerial imagery and odometry~\cite{mcconnell2022overhead}, while another used an orthographic imaging sonar within a submap-based framework to achieve dense 3-D reconstructions~\cite{mcconnell2024large}. Although these frameworks provide strong drift resilience and scalability, their reliance on landmark tracking and sparse or volumetric representations aligns them with navigation and 3-D reconstruction tasks rather than the creation of dense, textured 2-D mosaics.

In summary, direct registration and external anchoring methods offer lightweight pipelines better suited for constructing detailed 2-D overhead maps, whereas SLAM-centric approaches prioritize localization and structural mapping. Motivated by these distinctions, we develop a compact, drift-resilient 2-D seabed mapping framework that leverages robust local registration cues while incorporating GPS constraints, without resorting to full SLAM formulations.

\subsection{Challenges in ASV-Based Seabed Mapping}

Unlike AUVs, ASVs can directly access GPS for global positioning~\cite{karapetyan2025oysterbot}. Early surveys, such as the United States Geological Survey (USGS) studies in Long Island Sound~\cite{poppe2001}, registered sidescan sonar imagery to GPS tracks with manual corrections, illustrating both the utility and limitations of GPS anchoring. While GPS provides reliable global references, its spatial resolution is not enough for high-resolution mapping. On the other hand, sonar-based registration preserves local details but suffers from accumulated drift over long trajectories. Therefore, there is a critical need for methods that can help achieve drift-resilient, high-accuracy seabed mapping with ASVs to overcome the aforementioned constraints.

\subsection{Image Blending in Mapping}

Blending overlapping sonar images into a unified map is essential but often underexplored. In the broader field of image processing, image fusion techniques have been extensively surveyed~\cite{masood2017image}, which can be divided broadly into spatial-domain (e.g., averaging, principal component analysis (PCA), Brovey transforms) and transform-domain (e.g., pyramid- or wavelet-based) approaches. Spatial domain approaches are simple but prone to artifacts, while transform-based strategies generally preserve more detail and remain widely adopted across domains.  

For sonar-specific applications, Hurtós et al.~\cite{hurtos2013blending} proposed a blending pipeline tailored to FLS mosaicing that showed improved robustness to photometric irregularities and sonar-specific artifacts arising from the imaging geometry, although residual artifacts may still appear in overlapping regions.

Su et al.~\cite{SU2024117249} proposed a variance-based blending scheme that uses short- and long-term variance statistics to adaptively weight overlapping pixels to improve visual consistency in sonar mosaics. However, the robustness was limited due to the reliance on fixed temporal windows and simple averaging. Our approach is intended to enhance temporal modeling and fusion robustness, yielding smoother transitions and more reliable texture preservation.

As such, existing sonar mosaicing methods lacks drift
correction, GPS fusion alone is too coarse, and no prior work
integrates these into a lightweight, field-deployable ASV
pipeline.  Our framework integrates local FMT-based alignment, GPS/compass-constrained global optimization, and variance-based blending into a single pipeline specifically designed for low-cost ASVs operating in shallow, turbid waters. This framework enables accurate sub-meter level mapping compare to sonar-only or GPS-only approaches.
\section{APPROACH}
\label{section:approach}

We propose a drift-resilient global optimization framework for seabed mapping using FLS imagery. The framework integrates pairwise FLS-based local registrations with EKF-smoothed GPS/compass priors and refines them through nonlinear optimization to construct a globally consistent 2-D seabed maps. Our formulation employs a 2-D registration model, which is appropriate for near-planar seabed conditions. For seabeds with large slopes or complex 3-D topologies, mosaic quality may degrade. In this sections, we discuss the problem formulation followed by the detailed system architecture and components.

\subsection{Problem Formulation}

Our objective is to estimate globally consistent poses for a sequence of sonar images and fuse them into a single georeferenced mosaic. The inclusion of EKF-smoothed global priors reduces long-horizon drift and provides additional tolerance to small roll/pitch disturbances.

\textbf{Input:} A sequence of sonar images $\{I_1, I_2, \dots, I_n\}$ and corresponding ASV pose observations $\{G_1, G_2, \dots, G_n\}$, where each $G_i = (G_i^{(x,y)}, G_i^\theta)$ comprises a global position $G_i^{(x,y)}$ (from GPS) and a heading $G_i^\theta$ (from compass), both refined through EKF integration with IMU observations.

\textbf{Output:} Optimized global positions $\{P_1, P_2, \dots, P_n\}$, orientations $\{\Theta_1, \Theta_2, \dots, \Theta_n\}$, and a fused 2D map $\mathcal{M}$.  

Formally, let $\mathcal{P} = \{P_i, \Theta_i \mid i=1,\dots,n\}$ denote the global poses to be optimized. The optimization problem is:
\begin{equation}
\mathcal{P}^* = \arg\min_{\mathcal{P}} \mathcal{E}(\mathcal{P}),
\end{equation}

where $\mathcal{E}$ is a residual function combining consistency with local registration and agreement with global priors.

\begin{figure}[t]
\vspace{2mm}
    \centering
    \includegraphics[width=\columnwidth]{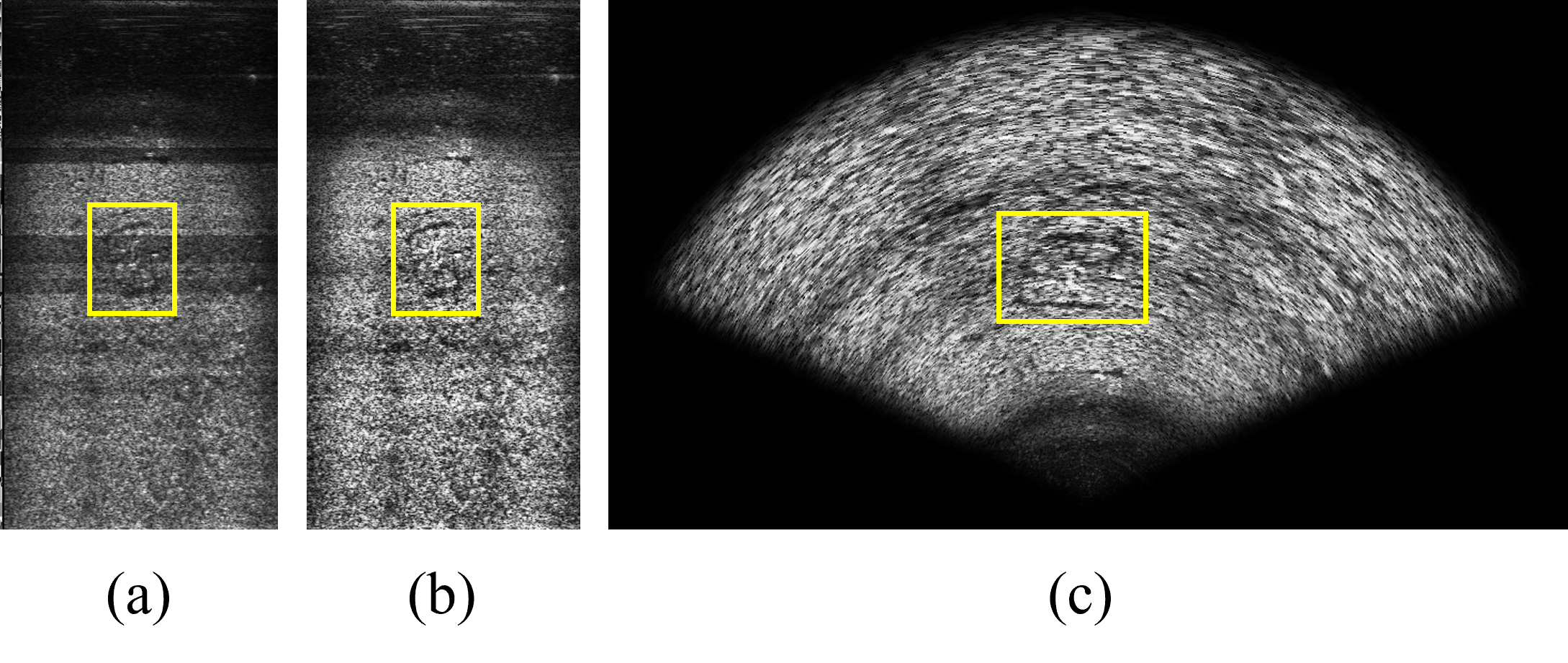}
    \vspace{-5mm}
    \caption{Comparison of raw and processed sonar images (5~m range, 130$^\circ$ field of view (FOV)): (a) and (b) are raw and processed images in polar coordinates, respectively. (c) is processed image in Cartesian coordinates. A quadrat (0.62~m side length) is enclosed by the yellow bounding box.}
    \label{fig:preproc}
\end{figure}

\begin{algorithm}[H]
\caption{\textsc{FMT-GPS-OPT with Robust Blending}}
\label{algorithm:fmt_gps_opt_blending}
\textbf{Input:} sonar images $\{I_1,\dots,I_n\}$, pose measurements $\{G_1,\dots,G_n\}$, scale $s$, thresholds $\tau_p,\tau_a$, weights $\lambda,\alpha$, EMA factors $\beta_s,\beta_l$, ratio $\rho$, Top-$N$, MAD factor $\eta$, exponent $\gamma$ \\
\textbf{Output:} 2D map $\mathcal{M}$

\begin{algorithmic}[1]

\State \textit{\textbf{Local Registration}}
\State $\mathcal{V}\gets\{1\}$ \Comment{valid frame indices}
\For{$i\gets 1$ \textbf{to} $n-1$}
    \State $(\Delta p_i,\Delta\theta_i)\gets \textsc{FMTRegistration}(I_i,I_{i+1}, s)$
    \State $(\delta_{p,i}, \delta_{\theta,i}) \gets \textsc{Val}((\Delta p_i,\Delta\theta_i),(\Delta p_{\mathrm{glob},i},\Delta\theta_{\mathrm{glob},i}))$

    \If{$\delta_{p,i} > \tau_p$ \textbf{or} $\delta_{\theta,i} > \tau_a$}
        \State discard $i{+}1$
    \Else
        \State $\mathcal{V}\gets\mathcal{V}\cup\{i{+}1\}$
    \EndIf
\EndFor

\State \textit{\textbf{Global Optimization}}
\State $\mathcal{R} \gets \textsc{GetResiduals}(\mathcal{P}, (\Delta p_i, \Delta \theta_i), (G_i^{(x,y)}, G_i^\theta))$
\State $\mathcal{P}^* \gets \textsc{Optimize}(\mathcal{R}; \lambda, \alpha)$

\State \textit{\textbf{Robust Blending}}
\For{$i\in\mathcal{V}$}
    \State Warp $I_i$ onto the canvas using $(\Theta_i,P_i)\in \mathcal{P}^*$
    \State $(v_{i,s},v_{i,l}) \gets \textsc{UpdateVarianceMaps}(I_i, \beta_s,\beta_l)$
    \State $s_i \gets \textsc{ComputeFeatureScore}(I_i, v_{i,s}, v_{i,l})$
\EndFor

\State Initialize $\mathcal{M}$
\For{each canvas pixel $(x,y)$}
    \State $\mathcal{K}'(x,y) \gets \textsc{SelectCandidates}(x,y,\rho,N, \eta)$
    \State $\mathcal{M}(x,y) \gets \textsc{RobustBlend}(\mathcal{K}'(x,y),\gamma)$
\EndFor

\State \Return $\mathcal{M}$
\end{algorithmic}
\label{alg:pipeline}
\end{algorithm}


\subsection{Architecture}

Our proposed system follows a modular architecture consisting of image preprocessing, local registration, global optimization, and blending (as shown in Fig.~\ref{fig:workflow}, also summarized in Algorithm~\ref{alg:pipeline}). To handle varying dataset scales, we implement three mapping strategies:

\textbf{Baseline:} Frames are aligned solely based on the \emph{Local Registration} block of Algorithm~\ref{alg:pipeline}, skipping \emph{Global Optimization} and proceeding directly to \emph{Robust Blending}. This configuration is used for the ablation study.

\textbf{Single-stage:} The complete pipeline of Algorithm~\ref{alg:pipeline} is executed in this design, which is effective for small to medium-scale datasets.

\textbf{Two-stage:} The sequence is divided into subsets with 50\% overlap, balancing computational cost and registration success. In Stage 1, each subset is processed using the \emph{Baseline} procedure. In Stage 2, the resulting submaps are aligned and fused using the complete pipeline of Algorithm~\ref{alg:pipeline}, identical to the \emph{Single-stage} setting. This hierarchical design enables better preservation of local details and reduces computational complexity for large-scale datasets.

\subsection{Preprocessing}

Prior to registration, the following preprocessing steps are performed to reduce sonar-specific artifacts and enhance local textures (Fig.~\ref{fig:preproc}):  
(1) Row-wise intensity equalization on polar frames to suppress radial ring artifacts via mean subtraction per scanline.  
(2) Optional median filtering with kernel size 5, or non-local means (NLM)~\cite{buades2005} filtering with filter strength $h=15$, template window size 7, and search window size 21, to remove background noise.  
(3) Global histogram matching with a reference image to normalize intensity, followed by contrast limited adaptive histogram equalization (CLAHE)~\cite{clahe1994} with clip limit 1.5 and tile grid size 16 to enhance local texture.  
(4) Optional Gaussian blurring with kernel size 3 and automatically determined standard deviation to prevent over-sharpening induced by CLAHE.  
(5) Normalization and conversion from polar-to-Cartesian coordinates, with edge smoothing using a Hamming window to suppress boundary discontinuities.    

\subsection{EKF-based Sensor Fusion for Enhanced Positioning}

    GPS receivers provide absolute localization but have limitations in accuracy and update rates. In contrast, IMUs deliver high-frequency measurements of angular rate and acceleration but suffer from drifts. To leverage the complementary strengths of these sensors, we employ an EKF that fuses GPS, IMU, and compass data to yield smooth, accurate estimates of 2-D position and heading.

The EKF maintains an 8-dimensional state vector
\begin{equation}
\mathbf{x}_k =
\begin{bmatrix}
p_{x},\,p_{y},\,v_{x},\,v_{y},\,\psi,\,b_{\omega},\,b_{a_x},\,b_{a_y}
\end{bmatrix}^\top,
\end{equation}
where $(p_x,p_y)$ are local positions, $(v_x,v_y)$ are velocities, $\psi$ is heading, and $b_\omega,b_{a_x},b_{a_y}$ represent IMU biases.

State propagation uses IMU angular rates and accelerations through a nonlinear kinematic model, while GPS provides direct position updates and the compass supplies heading observations. The corresponding measurement matrices are linear in the state. At each step, the EKF performs the standard \emph{predict} stage using IMU inputs, and when GPS or compass data are available, an \emph{update} is performed using the Kalman gain:
\begin{equation}
\hat{\mathbf{x}}_{k|k} = \hat{\mathbf{x}}_{k|k-1} + K_k\bigl(z_k - H \hat{\mathbf{x}}_{k|k-1}\bigr),
\end{equation}
with covariance updated accordingly.

After processing all $N$ measurements, the corrected positions and headings are sampled and linearly interpolated to provide a smooth trajectory estimate.

\subsection{Local Registration}

Pairwise registration between consecutive frames is performed using the FMT, a widely used phase correlation method~\cite{hurtos2012,hurtos2015}.  
For each consecutive pair $(I_i, I_{i+1})$, the relative displacement $\Delta p_i$ and rotation $\Delta \theta_i$ are estimated:
\begin{equation}
  (\Delta p_i, \Delta \theta_i) = \textsc{FMTRegistration}(I_i, I_{i+1}, s),  
\end{equation}
where $s$ is the pixel scale in meters.  

From global priors, relative motion is computed as
\begin{equation}
\Delta p_{\text{glob}, i} = G_{i+1}^{(x, y)} - G_i^{(x, y)}, \quad
\Delta \theta_{\text{glob}, i} = G_{i+1}^\theta - G_i^\theta.
\end{equation}
Motion differences relative to global priors are then evaluated:
\begin{equation}
\delta_{p,i} = \|\Delta p_i - \Delta p_{\text{glob}, i}\|, \quad
\delta_{\theta,i} = \big|\Delta \theta_i - \Delta \theta_{\text{glob}, i}\big|.
\end{equation}
The pair $(I_i, I_{i+1})$ is rejected if $\delta_{p,i} > \tau_p$ or $\delta_{\theta,i} > \tau_a$.
This procedure yields accurate local alignment while constraining drift through global priors, and corresponds to the \textit{Local Registration} block in Algorithm~\ref{alg:pipeline} (lines 1--11).

\subsection{Global Optimization}

The global optimization module refines the ASV trajectory by combining local registration residuals with global constraints. Specifically, for each consecutive pair, the local and global residuals, respectively, are defined as:
\begin{align}
\left\{
\begin{aligned}
r_{\text{local,pos}, i} &= (P_{i+1} - P_i) - \Delta p_i,\\
r_{\text{local,angle}, i} &= (\Theta_{i+1} - \Theta_i) - \Delta \theta_i. \\
r_{\text{glob,pos}, i} &= P_i - G_i^{(x,y)},\\
r_{\text{glob,angle}, i} &= \Theta_i - G_i^\theta. \label{eq:res_global_ang}
\end{aligned}
\right.
\end{align}

The optimization problem is then defined as:
\begin{align}
\mathcal{P}^* = \arg\min_{\mathcal{P}} \sum_i \Big(
   & \lambda \big( \|r_{\text{local,pos}, i}\|^2 
   + \alpha \|r_{\text{local,angle}, i}\|^2 \big) \notag \\
   & + \big( \|r_{\text{glob,pos}, i}\|^2 
   + \alpha \|r_{\text{glob,angle}, i}\|^2 \big) \Big).\label{eq:opt}
\end{align}

Solved via nonlinear least squares initialized from global measurements, in Algorithm~\ref{alg:pipeline}, these residuals are encapsulated by \textsc{GetResiduals} and optimized in the \textit{Global Optimization} block (lines 12--14).

\subsection{Image Blending}

After optimization, each image is rotated and aligned according to its refined orientation $\Theta_i$ and position $P_i$. To resolve conflicts in overlapping regions, we adopt a variance-based blending strategy derived from Su et al.~\cite{SU2024117249}, with modifications to improve robustness. For each image, a local variance map is computed, and both short-term and long-term statistics are smoothed using EMA. Specifically, for frame $i$, the EMA updates are defined as
\begin{align}
v_{i,s} &= \beta_s \, v_{i-1,s} + (1-\beta_s)\,\tilde{v}_{i}, \label{eq:ema_short}\\
v_{i,l} &= \beta_l \, v_{i-1,l} + (1-\beta_l)\,\tilde{v}_{i}, \label{eq:ema_long}
\end{align}
where $\tilde{v}_{i}$ denotes the instantaneous local variance map, and $\beta_s, \beta_l \in (0,1)$ are the decay factors for the short-term and long-term statistics, respectively. Larger $\beta$ values yield smoother but more inertial statistics, replacing fixed-size temporal windows and improving feature stability near sequence boundaries.

The feature score at pixel $(x,y)$ is then computed as
\begin{equation}
s_i = v_{i,s}^{\mathrm{norm}} \cdot e^{-v_{i,l}^{\mathrm{norm}}},
\label{eq:score}
\end{equation}
where $v_{i,s}$ and $v_{i,l}$ are normalized within the valid fan-shaped region. The resulting scores form the feature score map (FSM) for frame $i$. For each canvas pixel $(x,y)$, candidate intensities from overlapping frames are filtered by an adaptive threshold relative to the maximum score. Specifically, the retained set is defined as
\begin{equation}
\mathcal{K}(x,y)=\Big\{\,i:\ s_i(x,y)\ge \rho \cdot \max_j s_j(x,y)\,\Big\},
\end{equation}
where $\rho$ is the relative threshold parameter, and at most $N$ highest-scoring candidates are retained. To further suppress outliers, median absolute deviation (MAD)-based trimming with factor $\eta$ is applied to the retained intensity set, yielding a trimmed set $\mathcal{K}'(x,y)$. 

A weighted trimmed mean is then computed as
\begin{equation}
\mathcal{M}(x,y) = \frac{\sum_{i \in \mathcal{K}'(x,y)} w_i \, I_i}{\sum_{i \in \mathcal{K}'(x,y)} w_i}, 
\quad w_i = s_i^\gamma,
\label{eq:blending}
\end{equation}
where $\mathcal{M}(x,y)$ is the final blended intensity at canvas location $(x,y)$, $I_i$ is the intensity from the $i$-th overlapping frame, and $\gamma$ controls the sharpness of score weighting. This strategy ensures seamless transitions, preserves fine texture, and suppresses striping and outlier artifacts. These steps correspond to the \textit{Robust Blending} block in Algorithm~\ref{alg:pipeline} (lines 15--26).

\section{Experimental Results}
\label{section:results}

We conducted a series of field experiments with an ASV equipped with FLS to evaluate our mapping framework. This section describes the experimental setup, followed by the results from three complementary performance evaluations.

\subsection{Experimental Setup}
The ASV surveyed a test site with a pre-deployed quadrat array following lawnmower trajectories. To assess robustness under varying motion conditions, a total of 15 trials were conducted with different survey paths. The collected dataset provides a controlled yet diverse basis for evaluating our mapping framework against the known quadrat geometry.
\subsubsection{ASV Platform}
The ASV was built on the chassis of a pontoon boat and driven by BlueRobotics T200 thrusters. Navigation and control relied on a Pixhawk PX4 sensor suite with an M8N GPS module and a magnetic compass. The IMU provided angular rate and acceleration at 55~Hz, while GPS and compass supplied absolute position and heading at 2~Hz. An Oculus M1200d FLS was mounted at a $15^{\circ}$ grazing angle on the back of the ASV. The FLS ($130^{\circ}$ FOV and 5 m range) performed scans at 2.1~MHz and 10~FPS.

\subsubsection{Deployment Environment}
Field trials were conducted in 3-4 feet deep saltwater test site using a $2\times4$ array of square quadrats arranged in two parallel rows (see Fig.~\ref{fig:field_diagram}). One row consisted of aluminum frames with 17~in ($\approx$ 0.43~m) side length, and the other of PVC frames with 24.5~in ($\approx$ 0.62~m) side length. The two rows were separated by 60~in ($\approx$ 1.52~m) between their nearest edges, and quadrats within each row were spaced 1~m apart center-to-center. Oysters were distributed inside the frames to introduce texture variation. 

\subsubsection{Implementation and Runtime}
All modules were executed on a desktop workstation CPU (Intel(R) Core(TM) i7-14700K @ 3.40~GHz, 64~GB RAM). For a representative sequence of 200 frames, the average runtimes for frame-to-frame registration and validation, global trajectory refinement (nonlinear optimization), and warping/rendering and blending are 109.3~s, 12.7~s, and 104.8~s, respectively.

\subsection{Image Processing Ablation}

\begin{table}[t]
\vspace{1mm}
\centering
\caption{Ablation of preprocessing.}
\label{tab:ablation_preprocessing}
\setlength{\tabcolsep}{3.5pt}
\footnotesize
\begin{tabular}{cccc|cccccc}
\toprule
Eq & Filter & CLAHE & GB & Ent. & Var. & RMS & Ten. & VoL & SS\\
\midrule
\multicolumn{9}{l}{\emph{No CLAHE (GB = --)}}\\
-- & -- & -- & -- & 0.905 & 0.036 & 0.189 & 0.333 & 0.687 & \textbf{0.184}\\
\checkmark & -- & -- & -- & 0.921 & 0.035 & 0.187 & 0.358 & 0.743 & 0.178\\
\checkmark & Median & -- & -- & 0.864 & 0.035 & 0.188 & 0.099 & 0.060 & 0.143\\
\checkmark & NLM & -- & -- & 0.892 & 0.035 & 0.188 & 0.268 & 0.538 & 0.180\\
\midrule
\multicolumn{9}{l}{\emph{With CLAHE (GB = --)}}\\
-- & -- & \checkmark & -- & \textbf{0.966} & \textbf{0.056} & \textbf{0.237} & 0.742 & 1.550 & 0.174\\
\checkmark & -- & \checkmark & -- & 0.963 & 0.052 & 0.228 & \textbf{0.804} & \textbf{1.701} & 0.166\\
\checkmark & Median & \checkmark & -- & 0.953 & 0.048 & 0.218 & 0.180 & 0.108 & 0.140\\
\checkmark & NLM & \checkmark & -- & 0.955 & 0.055 & 0.234 & 0.631 & 1.281 & 0.172\\
\midrule
\multicolumn{9}{l}{\emph{GB effect (only when CLAHE = \checkmark)}}\\
\checkmark & -- & \checkmark & \checkmark & 0.941 & 0.044 & 0.210 & 0.308 & 0.154 & 0.178\\
\bottomrule
\end{tabular}
\end{table}

\begin{table}[t]
\centering
\caption{Relative-pose consistency across methods (evaluated against FMT-only and global-prior-only alignments).}
\label{tab:relative_pose_consistency}
\setlength{\tabcolsep}{5pt}
\begin{tabular}{l|cc|cc}
\toprule
\multirow{2}{*}{Method} & \multicolumn{2}{c|}{FMT only} & \multicolumn{2}{c}{Global prior only} \\
 & MAE (m) & RMSE (m) & MAE (m) & RMSE (m) \\
\midrule
ORB + RANSAC        & 0.316 & 1.173 & 0.304 & 1.161 \\
SIFT + RANSAC       & 1.058 & 2.807 & 1.048 & 2.800 \\
ECC maximization    & 0.069 & 0.106 & 0.066 & 0.085 \\
FMT only            & --    & --    & 0.042 & 0.064 \\
Global prior only            & 0.042 & 0.064 & --    & -- \\
Ours (single-stage) & \textbf{0.005} & \textbf{0.006} & \textbf{0.038} & \textbf{0.062} \\
\bottomrule
\end{tabular}
\vspace{-4mm}
\end{table}

Each preprocessing configuration was evaluated over 15 trials, 
with each trial consisting of more than 400 FLS images. 
The reported results are averaged across all trials, 
covering combinations of row‐wise brightness equalization, 
median filtering vs.\ NLM denoising, CLAHE contrast enhancement, 
and Gaussian blurring.

Representative before/after patches are shown in Fig.~\ref{fig:preproc}. The ablation results are summarized in Table~\ref{tab:ablation_preprocessing}. We report six indicators: entropy (Ent.)~\cite{shannon1948} reflecting information richness, 
variance (Var.) and RMS contrast (RMS)~\cite{peli1990} characterizing intensity variation, 
Tenengrad (Ten.)~\cite{krotkov1988} and variance of Laplacian (VoL)~\cite{pech2000} quantifying sharpness, 
and the similarity score (SS), defined as the peak of normalized cross-correlation (NCC) from FMT registration~\cite{reddy1996}, 
indicating registration consistency. 
The results show that CLAHE substantially enhances structural metrics, often achieving the highest scores, 
but it produces ring artifacts around bright regions that hinder blending. 
Combining row-wise equalization with CLAHE suppresses such artifacts and yields consistently strong performance across all metrics, providing the best option for addressing the trade-off between detail preservation and artifact suppression. Therefore, this approach is adopted as the most effective preprocessing scheme for the mapping pipeline.

\subsection{Registration Accuracy and Ablation}

Seven registration strategies were benchmarked:  
oriented FAST and rotated BRIEF (ORB)~\cite{rublee2011orb} + random sample consensus (RANSAC)~\cite{fischler1981random},  
scale-invariant feature transform (SIFT)~\cite{lowe2004distinctive} + RANSAC,  
enhanced correlation coefficient (ECC) maximization~\cite{evangelidis2008parametric},  
FMT-only,  
global-prior-only alignment,  
and our pipeline in both single- and two-stage variants.  
ORB and SIFT are classical feature detectors/descriptors, while RANSAC is a robust model fitting scheme. ECC maximization is an intensity-based (direct) parametric registration method that aligns images by maximizing the ECC.

\begin{figure}[t]
\vspace{2mm}
    \centering
    \begin{subfigure}[t]{0.32\linewidth}
        \centering
        \includegraphics[width=\linewidth]{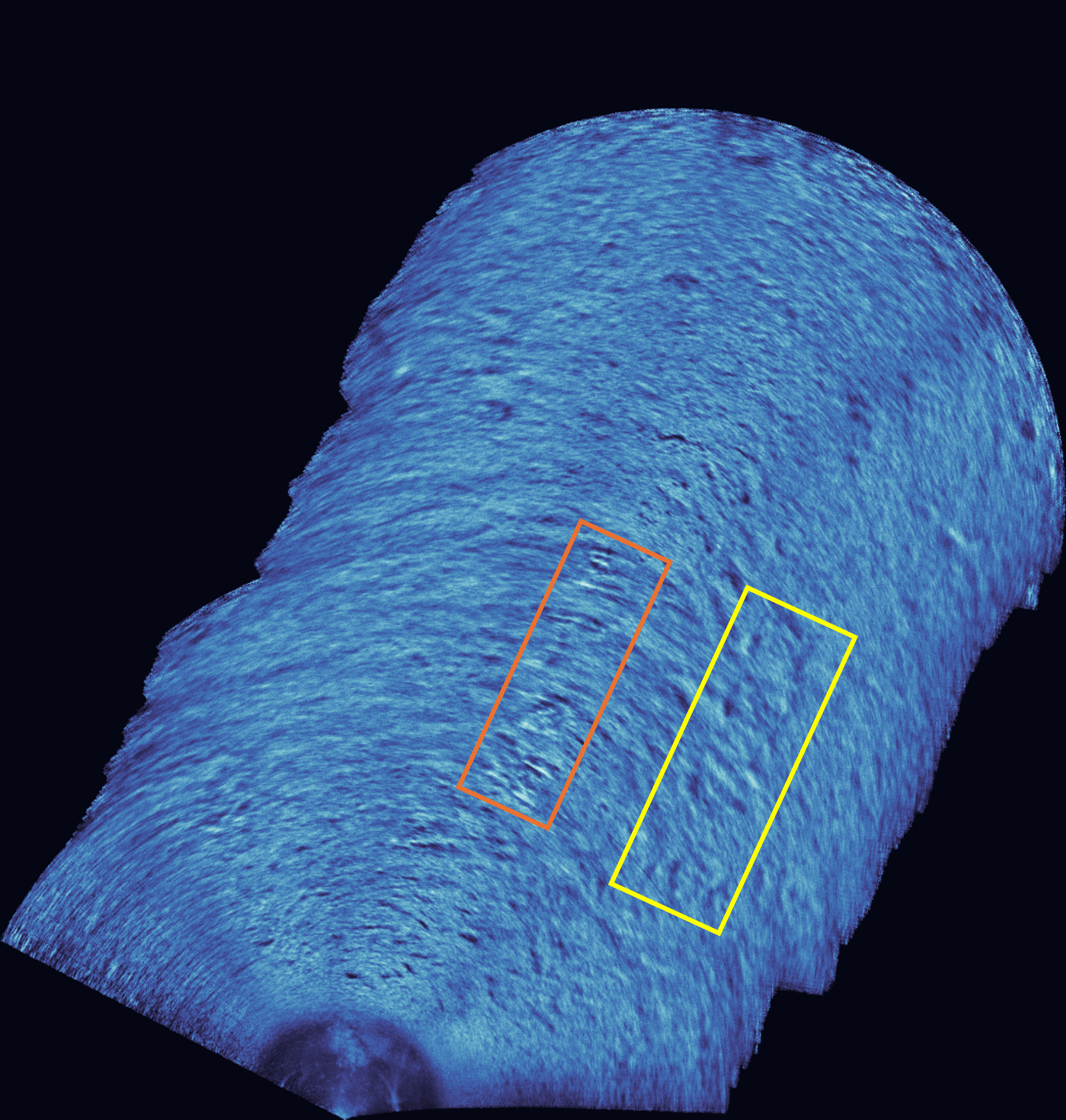}
        \caption{FMT-only}
        \label{fig:stage-fmt}
    \end{subfigure}
    \hfill
    \begin{subfigure}[t]{0.32\linewidth}
        \centering
        \includegraphics[width=\linewidth]{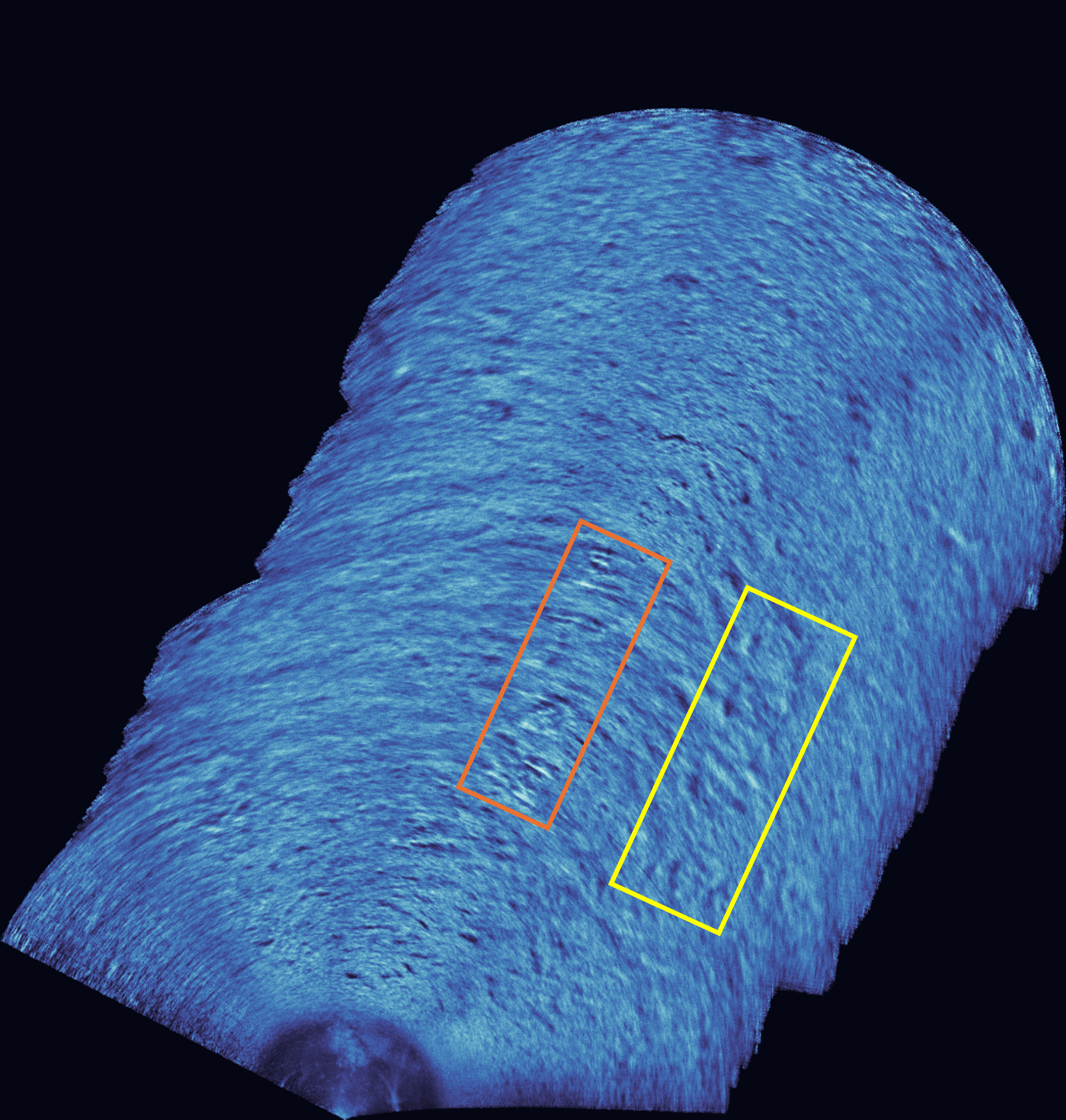}
        \caption{Single-stage}
        \label{fig:stage-single}
    \end{subfigure}
    \hfill
    \begin{subfigure}[t]{0.32\linewidth}
        \centering
        \includegraphics[width=\linewidth]{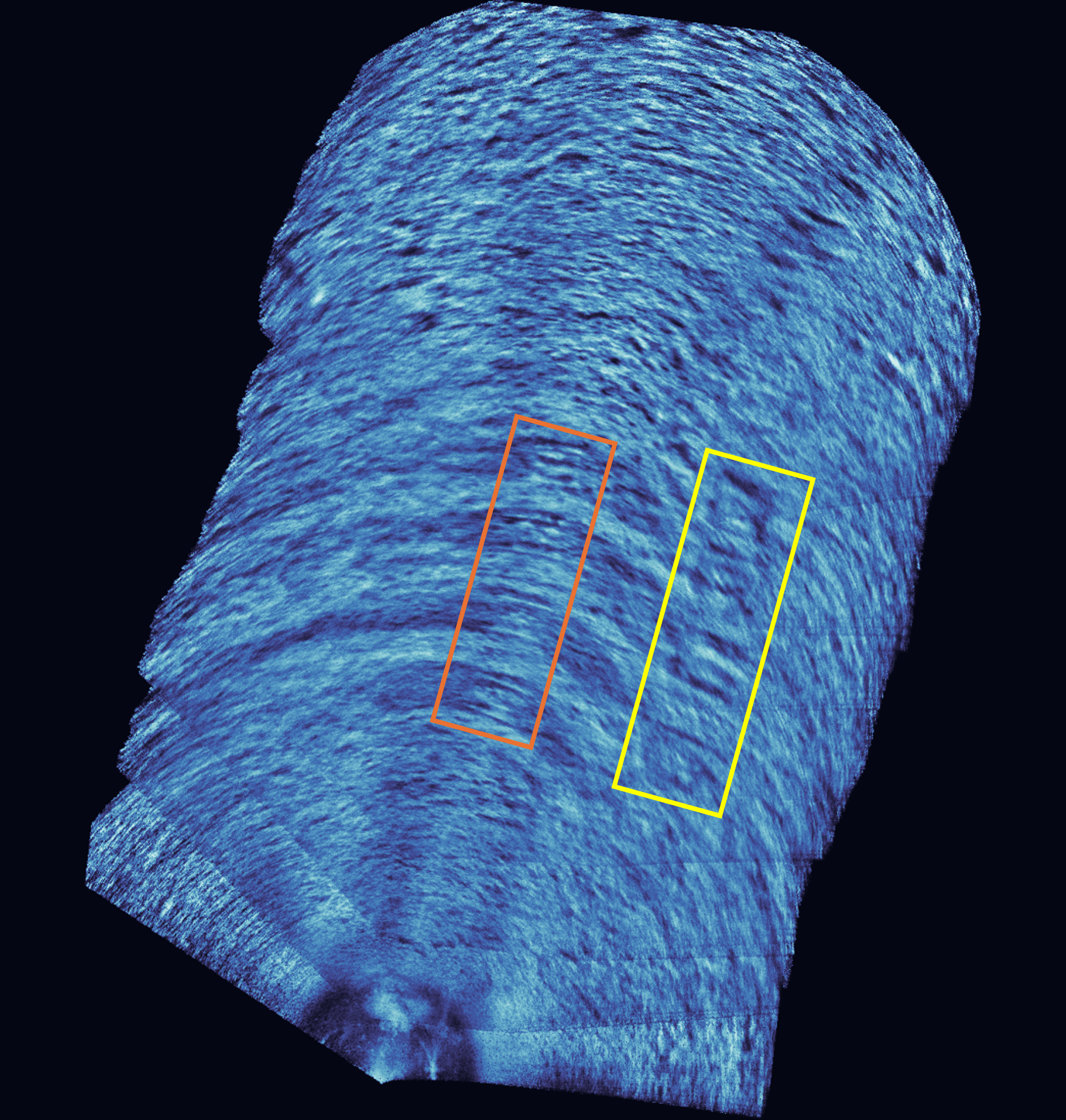}
        \caption{Two-stage}
        \label{fig:stage-double}
    \end{subfigure}

    \caption{Qualitative comparison of map registration on 180 FLS images, all using our proposed blending method. 
    (a) FMT-only baseline, map size 13.7\,m $\times$ 13.0\,m; 
    (b) single-stage, map size 12.6\,m $\times$ 14.1\,m; 
    (c) two-stage, map size 11.5\,m $\times$ 14.4\,m 
    (both with $\lambda=10^{4}$, $\alpha=\tfrac{1}{225}$).
    Orange and yellow bounding boxes enclose 4 aluminum and 4 PVC quadrats, respectively.}
    \label{fig:stage}
\end{figure}

 \begin{figure*}[t]
 \vspace{2mm}
     \centering
     \begin{subfigure}[t]{0.24\linewidth}
         \centering
         \includegraphics[width=\linewidth]{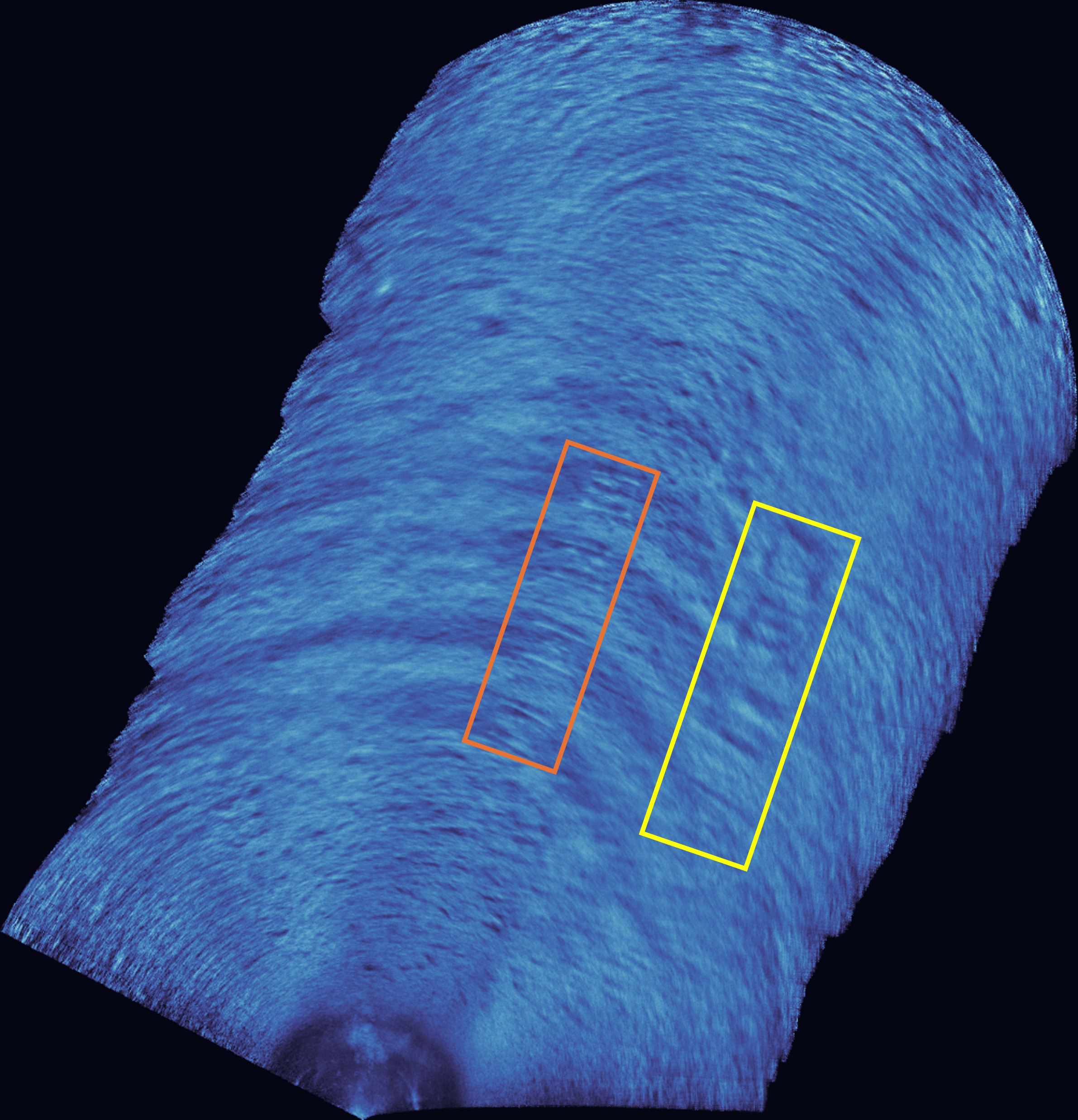}
         \caption{Averaging}
         \label{fig:blend_avg}
     \end{subfigure}
     \begin{subfigure}[t]{0.24\linewidth}
         \centering
         \includegraphics[width=\linewidth]{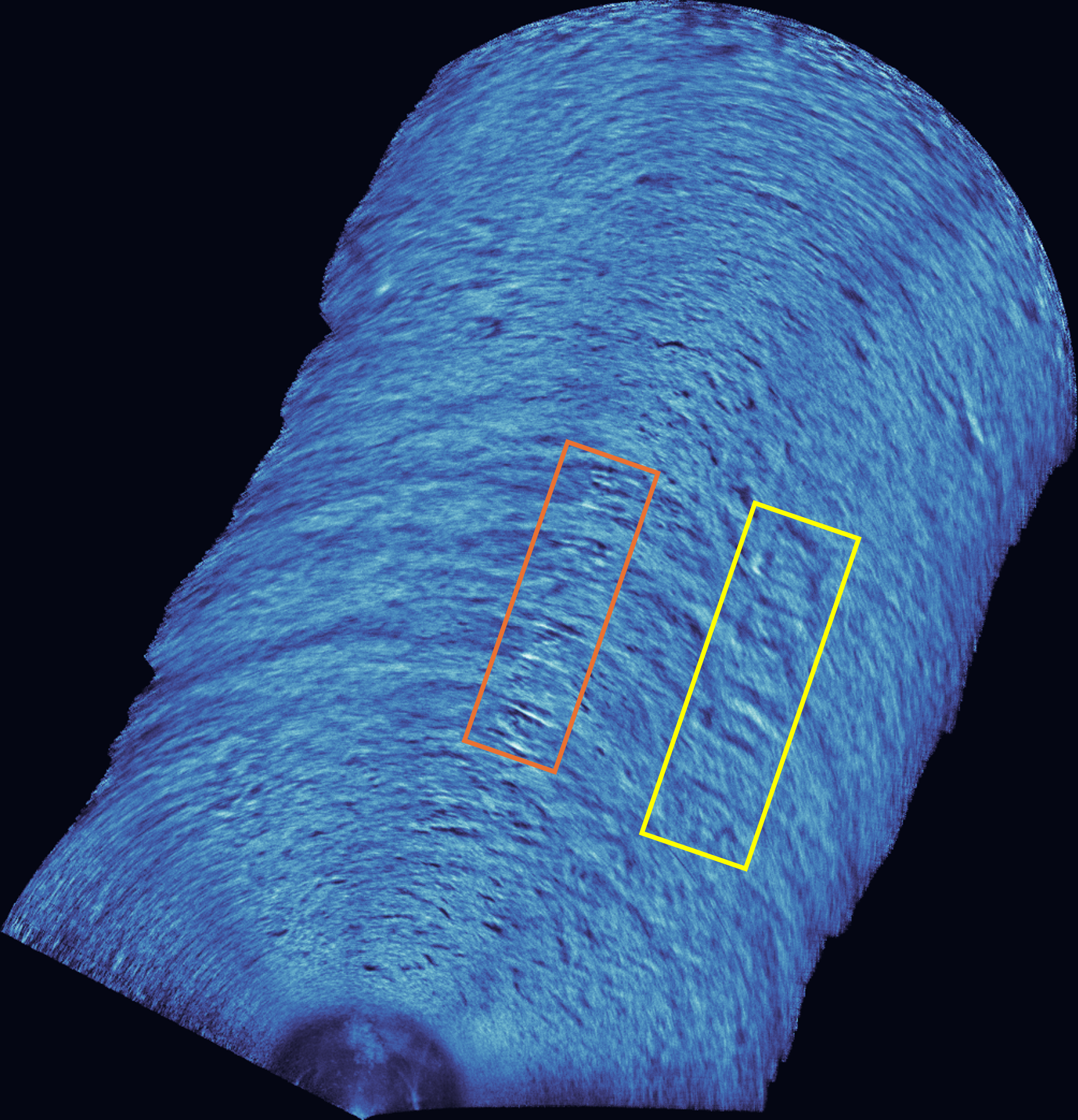}
         \caption{Su et al.\cite{SU2024117249}}
         \label{fig:blend_su}
     \end{subfigure}
     \hfill
     \begin{subfigure}[t]{0.24\linewidth}
         \centering
         \includegraphics[width=\linewidth]{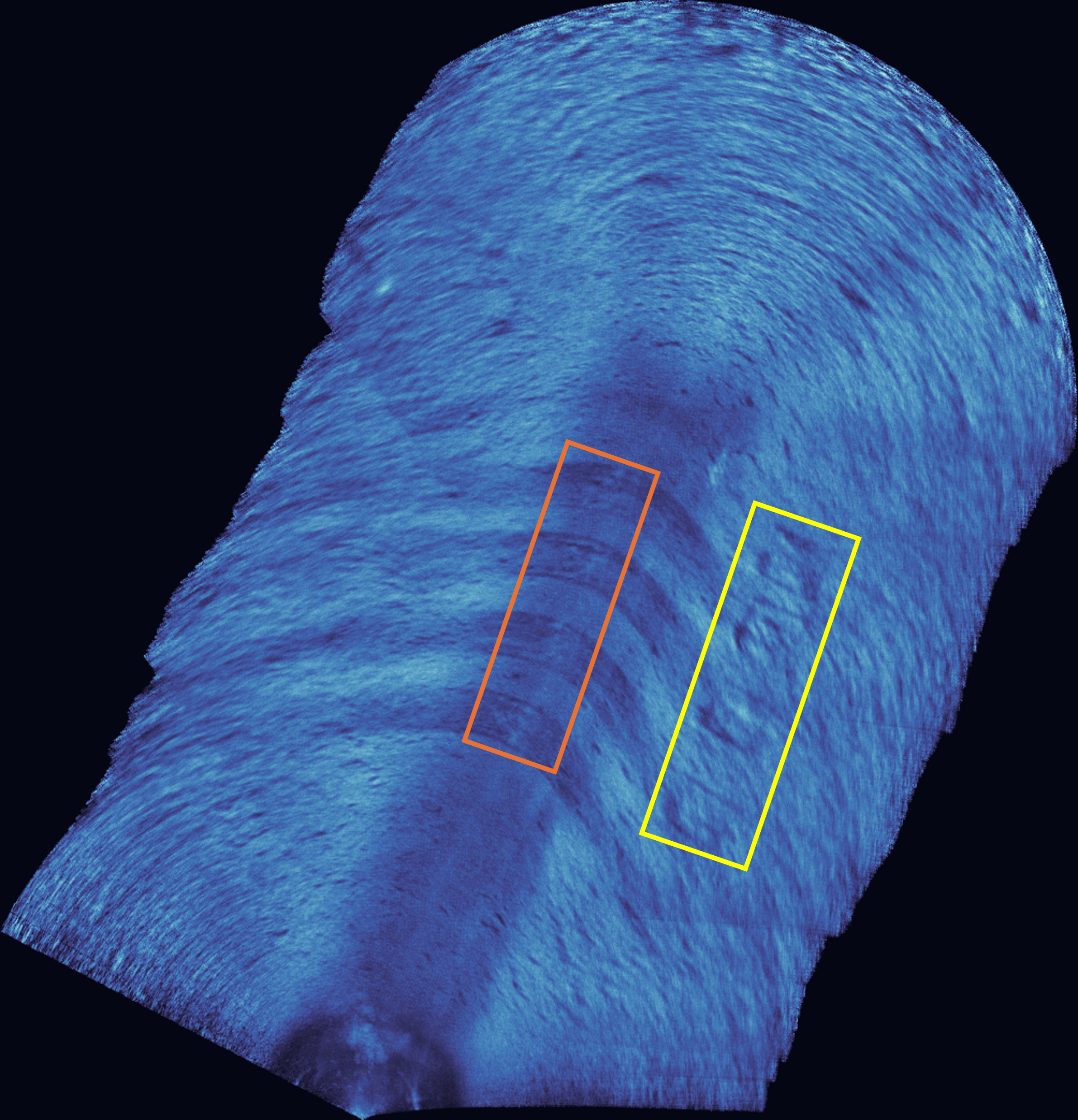}
         \caption{Hurtós et al.\cite{hurtos2013blending}}
         \label{fig:blend_hurtos}
     \end{subfigure}
     \hfill
     \begin{subfigure}[t]{0.24\linewidth}
         \centering
         \includegraphics[width=\linewidth]{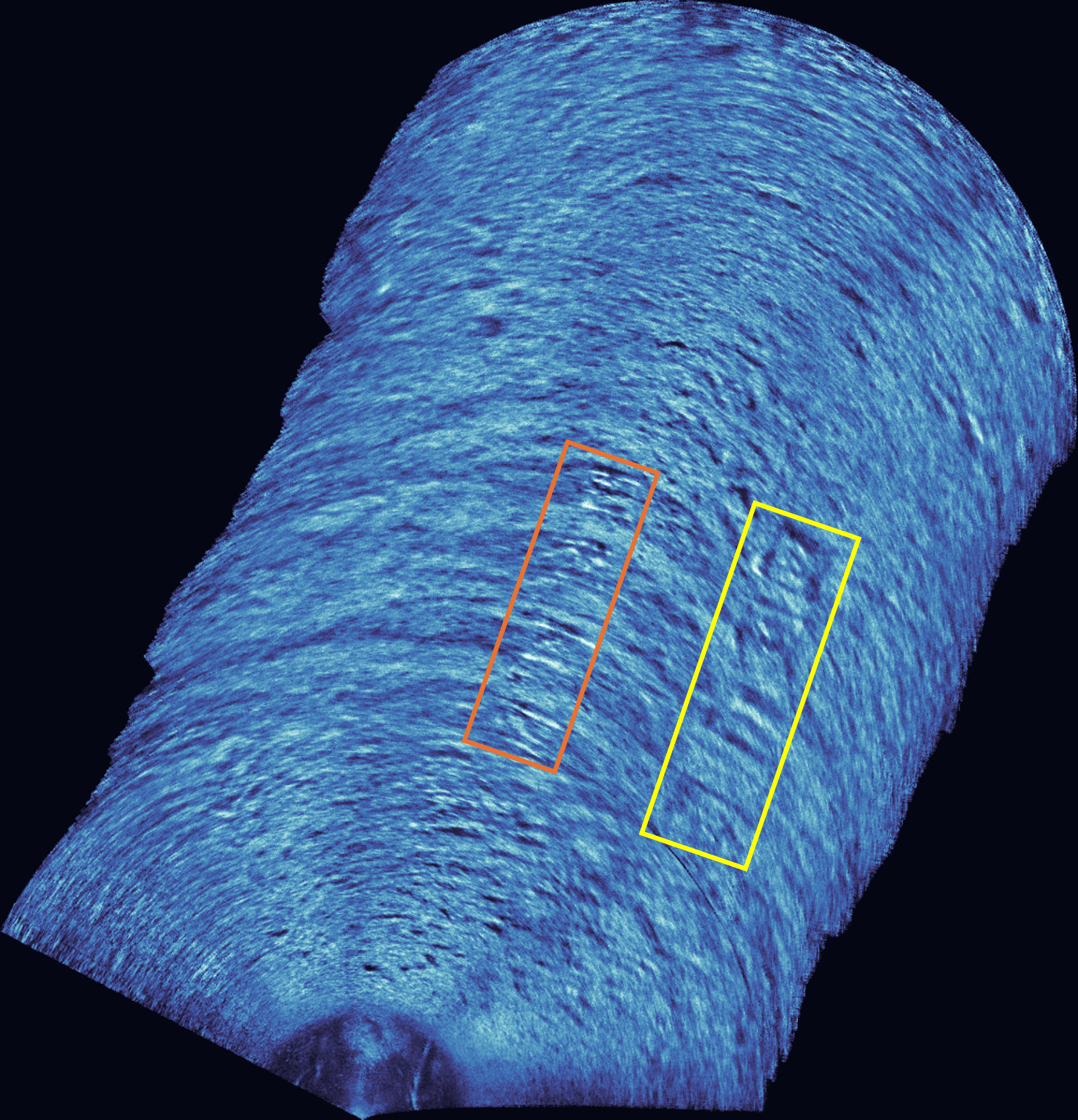}
         \caption{Proposed single-stage}
         \label{fig:blend_proposed}
     \end{subfigure}
     \hfill
     \caption{Qualitative comparison of map blending methods on a 13.2\,m $\times$ 13.7\,m area with 180 images, processed using the single-stage pipeline with $\lambda=10^{4}$ and $\alpha=\tfrac{1}{225}$:
     (a) simple intensity averaging,
     (b) variance‐based, 
     (c) distance‐based, and 
     (d) proposed blending approaches.
     Orange and yellow bounding boxes enclose 4 aluminum and 4 PVC quadrats, respectively.}
     \label{fig:blending}
     \vspace{-3mm}
 \end{figure*}

Table~\ref{tab:relative_pose_consistency} reports relative-pose consistency errors. Feature-based methods perform poorly on sonar imagery: ORB+RANSAC yields large errors, and SIFT+RANSAC fails outright with many outliers. ECC achieves low step-to-step errors but collapses on long sequences due to lacking reliable global constraints. Global-prior-only appears locally accurate but produces distorted maps with indistinguishable quadrats. FMT-only and our single-stage pipeline achieve strong local consistency, with the latter giving the lowest residuals. The two-stage pipeline is excluded here since trajectory-based consistency cannot be directly evaluated after global re-optimization.  

\begin{table}[t]
\centering
\caption{Quadrat spacing errors across methods.}
\label{tab:quadrat_spacing_errors}
\begin{tabular}{l|cc}
\toprule
Method           & MAE (m) & RMSE (m) \\
\midrule
FMT only            & 0.189 & 0.221 \\
Ours (single-stage) & \textbf{0.173} & \textbf{0.200} \\
Ours (two-stage)    & 0.176 & 0.207 \\
\bottomrule
\end{tabular}
\vspace{-3mm}
\end{table}

\begin{table}[t]
\centering
\caption{Effect of local registration weight $\lambda$ on quadrat spacing accuracy (with $\alpha=\tfrac{1}{225}$).}
\label{tab:lambda_ablation}
\begin{tabular}{c|cc}
\toprule
$\lambda$ & MAE (m) & RMSE (m) \\
\midrule            
$10^{2}$  & 0.476 & 0.606 \\
$10^{3}$  & 0.429 & 0.503 \\
$\mathbf{10^{4}}$  & \textbf{0.173} & \textbf{0.200} \\
$10^{5}$  & 0.201 & 0.248 \\
$10^{6}$  & 0.225 & 0.262 \\
\bottomrule
\end{tabular}
\vspace{-3mm}
\end{table}

Table~\ref{tab:quadrat_spacing_errors} evaluates quadrat spacing errors against the ground-truth $2 \times 4$ layout.  Our single-stage method performs the best. Compared with the FMT-only method, the MAE/RMSE errors reduce by 8.5\% and 9.5\%. The two-stage variant also outperforms the FMT-only method, with 6.9\% and 6.3\% reductions in MAE/RMSE errors. Global-prior-only is omitted here as it fails to recover quadrat geometry.

We further evaluated the effect of the local registration weight $\lambda$, which balances global constraints and frame-to-frame FMT alignment. The orientation-to-distance residual ratio was fixed at $\alpha=\tfrac{1}{225}$. Small values ($\lambda \leq 10^{1}$) were omitted since quadrats were not identifiable. Table~\ref{tab:lambda_ablation} shows that accuracy improves as $\lambda$ increases to $10^{4}$, at which the errors are lowest. Larger values ($10^{5}$–$10^{6}$) slightly degrade performance, indicating overfitting to noisy local estimates.

The three approaches are compared qualitatively in Fig.~\ref{fig:stage}. The single- and two-stage pipelines both reconstruct the global structure and the obtained quadrat spacing aligns well to the ground truth, while the FMT-only baseline, though retaining some local detail, shows evident drift. Between our two variants, the single-stage yields smoother stitching transitions and slightly lower spacing errors, whereas the two-stage strategy—by constructing overlapping submaps before global alignment—preserves sharper textures and clearer structural details. These results suggest that the single-stage pipeline is advantageous for seamless blending, while the two-stage pipeline enhances visual quality and robustness for longer sequences. Overall, our fusion methods achieve a balance of local consistency, global accuracy, and detail preservation, which outperforms the FMT-only baseline.

\subsection{Map Blending Performance}

\begin{table}[t]
\centering
\caption{Map blending quality metrics using different blending methods.}
\label{tab:blending_quality}
\begin{tabular}{l|ccccc}
\toprule
Method & Ent. & Var. & RMS & Ten. & VoL \\
\midrule
Averaging & 0.619  & 0.036 & 0.189 & 0.042 & 0.039 \\
Hurtós et al.  & 0.634 & 0.035 & 0.186 & 0.047 & 0.047 \\
Su et al.      & 0.656 & 0.043 & 0.208 & 0.084 & 0.077  \\
Ours     & \textbf{0.680} & \textbf{0.048} & \textbf{0.219} & \textbf{0.142} & \textbf{0.145} \\
\bottomrule
\end{tabular}
\vspace{-6mm}
\end{table}

We compared our proposed blending approach against three baselines under the single-stage pipeline: simple intensity averaging, the distance-based method of Hurtós et al.~\cite{hurtos2013blending}, and the variance-based method of Su et al.~\cite{SU2024117249}. For our method, the blending parameters were set to $\beta_s=0.60$, $\beta_l=0.97$, $\rho=0.68$, $N=16$, $\eta=2.7$, and $\gamma=1.4$.

As shown in Table~\ref{tab:blending_quality}, compared with Su et al., our method shows moderate improvements (3–12\%) in Ent., Var., and RMS, while Ten. and VoL are nearly doubled (from 0.084/0.077 to 0.142/0.145). 
Relative to simple averaging, Ten. and VoL increase more than threefold, highlighting substantial gains in edge sharpness and local detail.  Fig.~\ref{fig:blending} illustrates these differences. 
Our approach best preserves structural detail, ensures smooth transitions, and suppresses noise and artifacts. 
Su’s method is less sharp, Hurtós’s method suffers from severe blind-region artifacts, and averaging produces blurred, indistinct structures.  

Overall, through both quantitative and qualitative evaluations, our method is confirmed to achieve the most effective blending, offering improved balance between artifact suppression and detail preservation.

\section{CONCLUSION}
\label{section:conclusions}

We introduced a drift-resilient seabed mapping framework that couples local sonar registration, globally consistent trajectory refinement, and detail-preserving blending. By integrating GPS-based priors with refined FMT alignment, the system reduces long-term drift and attains sub-meter georeferenced accuracy, with a 9.5\% RMSE improvement over the FMT-only baseline and consistently better performance than representative registration baselines. The proposed variance-driven blending further enhances map quality by suppressing artifacts and preserving fine textures, yielding clearer mosaics than existing blending methods. Field trials in structured coastal environments confirm the practicality of this low-cost ASV-based solution for large-area autonomous seabed surveys.

\textbf{Limitations and Future Work:} While the proposed system produces accurate quadrat-level mosaics, the current formulation assumes near-planar seabed conditions with 2-D registration, which may degrade under strong 3-D relief or large slope conditions. In addition, the current implementation requires offline processing. As next steps, we will extend quadrat-level mapping to inventory surveys and perform substrate classification within each quadrat to estimate oyster distribution, size, and condition. Future work will also enable onboard online mapping to support informative path planning and efficient surveys of regions of interest. 

\vspace{-2mm}

%
\IEEEpeerreviewmaketitle




\bibliographystyle{IEEEtran}
\bibliography{references}

\end{document}